%% file: main.tex
\documentclass{article}

\usepackage[final,nonatbib]{nips_2016}
\usepackage{times}
\usepackage{epsfig}
\usepackage{graphicx}
\usepackage{amsmath}
\usepackage{amssymb}

\usepackage[utf8]{inputenc} 
\usepackage[T1]{fontenc}    
\usepackage{hyperref}       
\usepackage{url}            
\usepackage{booktabs}       
\usepackage{amsfonts}       
\usepackage{nicefrac}       
\usepackage{microtype}      
\usepackage{subcaption}
\usepackage{wrapfig}
\usepackage[font=scriptsize,labelfont=scriptsize]{caption}
\usepackage{array}

\usepackage{hyperref}
\usepackage{mysymbols}
\usepackage{cleveref}
\usepackage{float}
\usepackage{booktabs}       
\usepackage{comment}

\title{Grad-CAM: Why did you say that? 
}

\newcommand{\gb}{Guided Backpropagation}
\newcommand{\cgb}{Guided Grad-CAM}

\newcommand{\gcam}{Grad-CAM}

\newcommand{\reffig}[1]{Fig.~\ref{#1}}

\newcommand{\rp}[1]{\textcolor{black}{#1}}

\newcommand{\review}[1]{\textcolor{black}{#1}}

\newcommand{\what}[1]{\emph{what}}
\newcommand{\where}[1]{\emph{where}}
\newcommand{\whatcent}[1]{\emph{what}-centric}
\newcommand{\wherecent}[1]{\emph{where}-centric}
\newcommand{\Whatcent}[1]{\emph{What}-centric}
\newcommand{\Wherecent}[1]{\emph{Where}-centric}
\newcommand{\para}[1]{\textbf{#1}.}

\author{
	Ramprasaath R. Selvaraju \hspace{1.5pc}
	Abhishek Das \hspace{1.5pc}
	Ramakrishna Vedantam \hspace{1.5pc}
	Michael Cogswell \\
    \textbf{Devi Parikh \hspace{2pc}
	Dhruv Batra} \\
	Virginia Tech \\
    {\tt\small \{ram21, abhshkdz, vrama91, cogswell, parikh, dbatra\}@vt.edu}
\vspace{-10pt}
}

\begin{document}

\maketitle

\begin{abstract}
\input{sections/abstract}
\end{abstract}

\input{sections/intro}
\vspace{-12pt}
\input{sections/approach}

\vspace{-8pt}
\input{sections/human_evaluation}

\vspace{-20pt}
\input{sections/image_classification}
\input{sections/experiments}

\vspace{-5pt}
\input{sections/conclusions}
\vspace{-5pt}

\vspace{10pt}

{\small
\bibliographystyle{ieee}
\bibliography{strings,main}
}

\end{document}

%% file: sections/abstract.tex
\label{sec:abstract}
We propose a technique for making Convolutional Neural Network (CNN)-based models more transparent 
by visualizing input regions that are `important' for predictions -- producing \emph{visual explanations}.
Our approach, called Gradient-weighted Class Activation Mapping (Grad-CAM), uses class-specific gradient information 
to localize important regions.
These localizations are combined with existing
pixel-space visualizations to create a novel high-resolution and class-discriminative
visualization called \cgb{}.
These methods help better understand CNN-based models, including image
captioning and visual question answering (VQA) models. 
We evaluate our visual explanations by measuring their ability to discriminate between
classes, to inspire trust in humans, and their correlation with occlusion maps.
\gcam{} provides a new way to understand CNN-based models.

We have released code, an online demo hosted on CloudCV~\cite{agrawal2015cloudcv}, and the full paper~\cite{gradcam_arxiv}.\footnote{
code: \url{https://github.com/ramprs/grad-cam/}
demo: \url{http://gradcam.cloudcv.org}
paper: \url{https://arxiv.org/abs/1610.02391}
}

%% file: sections/intro.tex
\vspace{-5pt}
\section{Introduction}
\vspace{-10pt}

Convolutional Neural Networks (CNNs) and other deep networks have enabled unprecedented breakthroughs in a variety of computer vision tasks, from image classification
to image captioning 
and visual question answering. 
While these deep neural networks enable superior performance, their lack of decomposability into \emph{intuitive and understandable} components makes them hard to interpret. 
Consequently, when
today's intelligent systems fail, they fail spectacularly disgracefully,
without warning or explanation, leaving a user staring at incoherent output, wondering why the system did what it did.  
In order to build trust in intellegent systems and move towards their meaningful integration into our everyday lives,
it is clear that we must build `transparent' models that explain \emph{why they predict what they do}.

However, there is a trade-off between accuracy and simplicity/interpretability.
Classical rule-based or expert systems 
were highly interpretable but not very accurate (or robust).
Decomposable pipelines where each stage is hand-designed are thought to be more interpretable as each individual component assumes a natural intuitive explanation.
This tradeoff is realized in more recent work, like Class Activation Mapping (CAM)~\cite{zhou_cvpr16},
which allows explanations for a specific class of image classification CNNs.
By using deep models, we sacrifice a degree of interpretability in pipeline modules in order to achieve greater performance through greater abstraction (more layers) and tighter integration (end-to-end training).

\textbf{What makes a good visual explanation?}
Consider image classification -- a `good' visual explanation from the model justifying a predicted class should be (a) class-discriminative (\ie localize the category in the image) and (b) high-resolution (\ie capture fine-grained detail).
\review{To be concrete we introduce \gcam{} using the notion `class' from image classification
(\eg, cat or dog), but visual explanations can be considered
for any differentiable node in a computational graph, including words
from a caption or the answer to a question.}

\begin{figure}
    \begin{center}
  \centering
  \includegraphics[width=\linewidth]{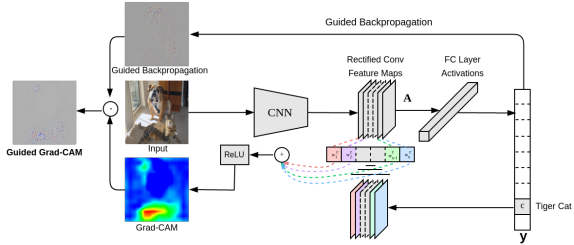}
    \vspace{-18pt}
	\caption{\scriptsize{\gcam{} overview: Given an image, and a category (`tiger cat') as input, we foward propagate the image through the model to obtain the raw class scores before softmax.
        The gradients are set to zero for all classes except the desired class (tiger cat), which is set to 1.
	This signal is then backpropagated to the rectified convolutional feature map of interest, where we can compute the coarse \gcam{} localization (blue heatmap).
	Finally, we pointwise multiply the heatmap with guided backpropagation to get \cgb{} visualizations which are both high-resolution and class-discriminative.
}}
	\label{fig:approach}
\end{center}
\vspace{-30pt}
\end{figure}

This is illustrated in \reffig{fig:approach}, where we visualize the `tiger cat' class.
Pixel-space gradient visualizations such as \gb{}~\cite{springenberg_arxiv14}, seen at the top of \reffig{fig:approach}, are high-resolution and highlight fine-grained details in the image, but are not class-discriminative.
Both the cat and the dog are highlighted despite `tiger cat' being the class of interest (in fact, the \gb{} visualizations of `boxer' dog `tiger cat' are indistinguishable).
However, the Grad-CAM visualization is low-resolution and does not contain fine details.
A high-res visualization like Guided Grad-CAM helps show these details by highlighting stripes in the cat in addition to localizing the cat.
We combine the best of both worlds by fusing existing pixel-space gradient visualizations with our novel localization method -- called \gcam{} -- to create \cgb{} visualizations, which are both high-resolution and class-discriminative.

In this abstract:
\textbf{(1)}
We propose Gradient-weighted Class Activation Mapping (\gcam{}) to generate visual explanations from \emph{any} CNN-based network without requiring architectural changes.
\textbf{(2)}
To illustrate the broad applicability of our technique across tasks, we apply \gcam{} to state-of-the-art 
image captioning and visual question answering models. 
\textbf{(3)}
We design and conduct human studies to show that \cgb{} explanations are class-discriminative and help humans not only establish trust, but also
help untrained users
successfully discern a `stronger' deep network from a `weaker' one
\emph{even when both networks make identical predictions, simply on the basis of their visual explanations.}
\textbf{(4)}
Our code and demos of \gcam{} are released to help others apply \gcam{} to interpret their own models.

%% file: sections/approach.tex
\section{Approach}\label{sec:approach}
\vspace{-6pt}
We review Class Activation Mapping (CAM)~\cite{zhou_cvpr16}, propose \gcam{},
and combine \gcam{} with high-resolution visualizations to form \cgb{}. 
This is summarized by \reffig{fig:approach}.

\para{Class Activation Mapping (CAM)}
CAM~\cite{zhou_cvpr16} produces a localization map from image classification 
CNNs where global-average-pooled convolutional feature maps are fed directly into a softmax. 
Specifically, let the penultimate CNN layer produce $K$ feature maps
$A^k \in \mathbb{R}^{u \times v}$ of width $u$ and height $v$.
These feature maps are then spatially pooled using Global Average Pooling (GAP)
and linearly transformed to produce a score $y^c$ for each class $c$
\vspace{-4pt}
\begin{equation} \label{eq:scores}
\vspace{-6pt}
    y_c = \sum_{k}
        \vphantom{\frac{1}{Z}\sum_i \sum_j} w^c_k
        \frac{1}{Z}\sum_i \sum_j
        \vphantom{\frac{1}{Z}\sum_i \sum_j} A^k_{ij}.
\vspace{-6pt}
\end{equation}
To produce a localization map $L_{\text{CAM}}^c \in \mathbb{R}^{u \times v}$ for class $c$, CAM computes the linear combination of the final
feature maps using the learned weights of the final layer:
$
    L_{\text{CAM}}^c = \sum_k w_k^c A^k.
$
This is normalized to lie between 0 and 1 for visualization purposes.
CAM can not be applied to networks which use multiple fully-connected layers before the output
layer, so fully-connected layers are replaced with convolutional ones and the
network is re-trained. In comparison, our approach can be applied directly to any CNN-based differentiable
architecture as is without re-training.



\para{Gradient-weighted Class Activation Mapping}
In order to obtain the class-discriminative localization map \gcam{} $L_{\text{\gcam{}}}^c \in \mathbb{R}^{u \times v} $ in generic CNN-based architectures, we
first compute the gradient of $y^c$ with respect to feature maps $A$ of a convolutional layer, \ie $\frac{\del y^c}{\del A^k_{ij}}$. These gradients are global-average-pooled to obtain weights $\alpha{}_{k}^c$:
\begin{equation} \label{eq:alpha}
\vspace{-6pt}
    \alpha{}_{k}^c =
        \frac{1}{Z}\sum_{i}\sum_{j}
        \vphantom{\sum_{i}\sum_{j}} \frac{\partial y^c}{\partial A_{ij}^{k}}
\vspace{-4pt}
\end{equation}
This weight $\alpha{}_{k}^c$ represents a \emph{partial linearization} of the deep network downstream from A, 
and captures the `importance' of feature map $k$ for a target class $c$.
In general, $y^c$ need not be a class score, but could be any differentiable activation.
As in CAM, our \gcam{} heat-map is a weighted combination of feature maps, but
we follow this by a ReLU:

\vspace{-13pt}
\begin{equation} \label{eq:gcam}
    L_{\text{\gcam{}}}^{c} = ReLU \left(\sum_k \alpha{}_{k}^{c} A^{k}\right)
\end{equation}
This results in a coarse heat-map, which is normalized for visualization. 
Other than the ReLU in \eqref{eq:gcam}, \emph{\gcam{} is a 
generalization of CAM} ($w_{k}^c$ are precisely $\alpha{}_{k}^{c}$ where CAM can be applied) \rp{to any CNN-based architecture (CNNs with fully-connected-layers, ResNets, CNNs stacked with Recurrent Neural Networks (RNNs) \etc.).}

\para{\cgb{}}
In order to combine the class-discriminative nature of \gcam{} and the high-resolution nature of \gb{}, we fuse them via point-wise multiplication
to create \cgb{}, shown on the left of \figref{fig:approach}.
We expect the last convolutional layers to have the best compromise between high-level semantics and detailed spatial information,
so we use these feature maps to compute \gcam{} and \cgb{}.


%% file: sections/human_evaluation.tex
\vspace{-5pt}
\section{Experiments}
\vspace{-10pt}
We evaluate our visualization then show image captioning and visual question answering examples.
\vspace{-7pt}
\subsection{Evaluating Visualizations} \label{sec:human_evaluation}
\vspace{-5pt}

\para{Evaluating Class Discrimination} \label{sec:class_disc}
Intuitively, a good prediction explanation is one that produces discriminative visualizations for the class of interest.
We select images from PASCAL VOC 2007 val set that contain exactly two annotated categories,
and create visualizations for one of the classes.
These are shown to workers on Amazon Mechanical Turk (AMT), who are asked ``Which of the two object categories is depicted in the image?''
and presented with the two categories present in the original image as options.
\review{As shown in \tabref{tab:eval_vis} column 1, human subjects can correctly identify the category being visualized substantially more often
when using \gcam{}.
This makes \gb{} more class-discriminative.}

\para{Evaluating Trust}
Given explanations from two different models, we want to evaluate which of them seems more trustworthy.
We use AlexNet and VGG-16 to compare \gb{} and \cgb{} visualizations, noting
that VGG-16 is known to be more reliable than AlexNet. 
In order to tease apart the efficacy of the visualization from the accuracy of the model being visualized, we consider only those instances where \emph{both} models made the same prediction as ground truth. 
Given a visualization from AlexNet, one from VGG-16, and the name of the object predicted by both the networks, workers are instructed to rate which model is more reliable. 
\review{Results are shown in the Relative Reliability column of \tabref{tab:eval_vis}, where scores range from -2 to +2 and positive scores indicate VGG
is judged to be more reliable than AlexNet.}
With \gb{}, humans score VGG as slightly more reliable than AlexNet, while with \cgb{} they score VGG as clearly more reliable than AlexNet.
Thus our \cgb{} visualization can help users place trust in a model that can generalize better, based on individual prediction explanations.

\para{Faithfulness \vs Interpretability} \label{ssec:occ}
Faithfulness of a visualization to a model is its ability to accurately explain the function learned by the model.
Naturally, there exists a tradeoff between the interpretability and faithfulness for complex models operating on highly compositional inputs.
A more faithful visualization might describe the model in precise detail yet be completely opaque to human inspection.
Here we are only interested in local fidelity; the visualization need only explain the parts of the model relevant to that image.
That is, in the vicinity of the input data point, our explanation should be faithful to the model~\cite{lime_sigkdd16}.

For comparison, we need a reference explanation with high local-faithfulness.
One obvious choice for such a visualization is image occlusion~\cite{zeiler_eccv14}, where we measure the difference in CNN scores when patches of the input image are masked out.
Interestingly, patches which change the CNN score are also patches to which \cgb{} assigns high intensity, \review{
as shown by measuring rank correlation between patch intensities in \tabref{tab:eval_vis} (3rd column).}
This shows that \cgb{} is more faithful to the original model than \gb{}.
\vspace{-5pt}
\begin{table}[h!]
\vspace{-2pt}
\centering
\resizebox{\columnwidth}{!}{%
    \begin{tabular}{c p{3cm} p{2.0cm} p{3.0cm}}\toprule
        \textbf{Method} & \textbf{Human Classification Accuracy} & \textbf{Relative Reliability} & \textbf{Rank Correlation \;\;\; w/ Occlusion} \\
        \midrule
        \gb{}  & 44.44 & +1.00 & 0.168 \\
        \cgb{} & 61.23 & +1.27 & 0.261 \\
        \bottomrule
    \end{tabular}}
    \caption{\scriptsize{\review{Quantitative Visualization Evaluation.
\cgb{} enables humans to differentiate between visualizations of different classes (Human Classification Accuracy) and
pick more reliable models (Relative Reliability). It also accurately reflects the behavior of the model (Rank Correlation w/ Occlusion).}}}
\label{tab:eval_vis}
   \vspace{-20pt}
\end{table}

%% file: sections/image_classification.tex
\subsection{Analyzing Failure Modes for VGG-16}
\vspace{-10pt}

\begin{figure}[ht!]
    \begin{center}
    \begin{subfigure}[b]{0.23\linewidth}
        \centering
        \includegraphics[width=1\linewidth]{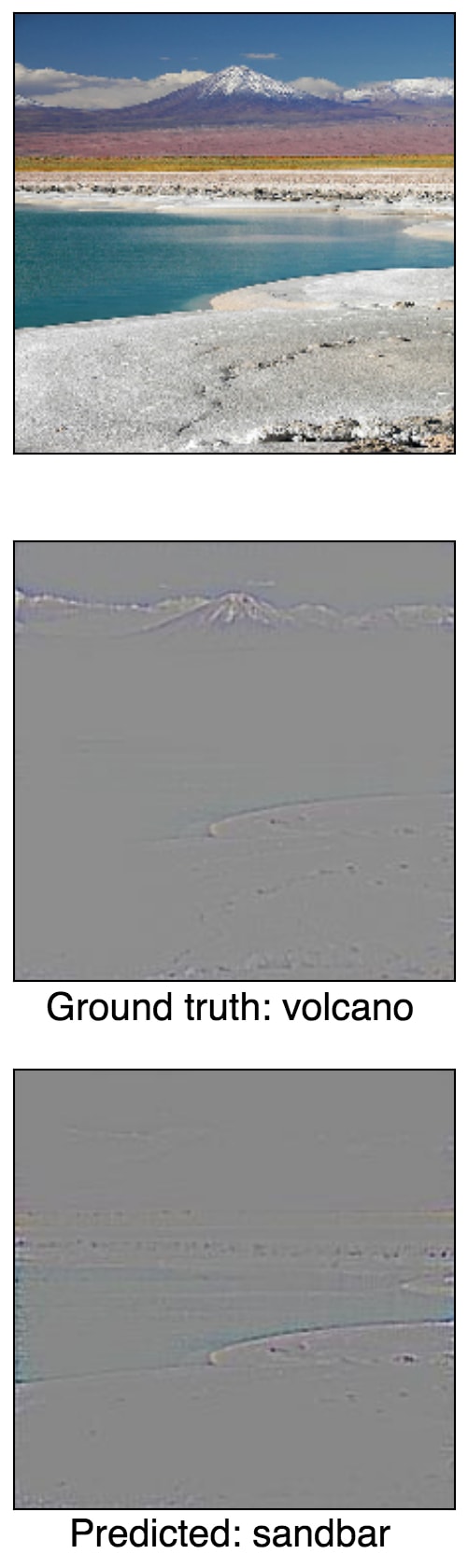}
        \vspace{-15pt}
        \caption{}
        \label{fig:failure_volcano}
    \end{subfigure}
    \begin{subfigure}[b]{0.23\linewidth}
        \centering
        \includegraphics[width=1\linewidth]{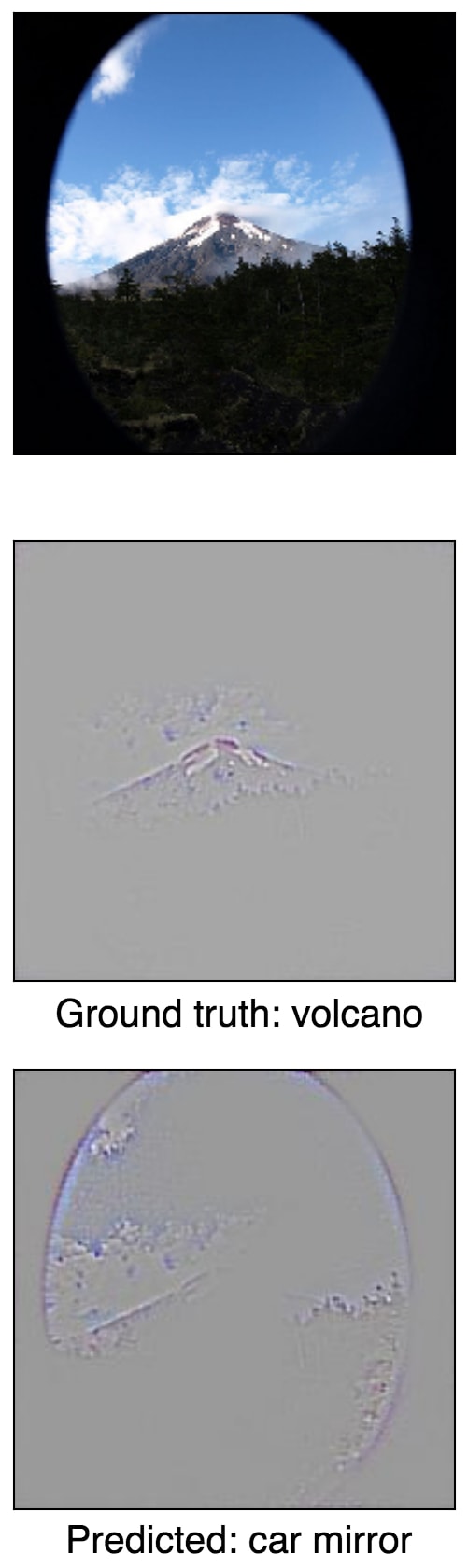}
        \vspace{-15pt}
        \caption{}
        \label{fig:failure_mirror}
    \end{subfigure}
    \begin{subfigure}[b]{0.23\linewidth}
        \centering
        \includegraphics[width=1\linewidth]{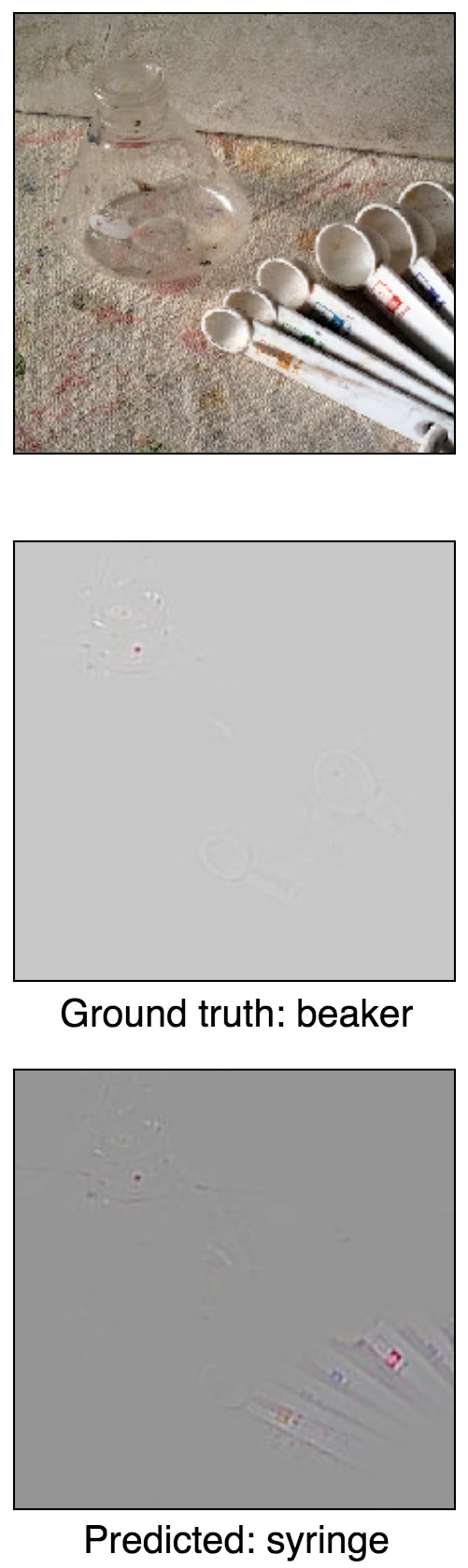}
        \vspace{-15pt}
        \caption{}
        \label{fig:failure_syringe}
    \end{subfigure}
    \begin{subfigure}[b]{0.23\linewidth}
        \centering
        \includegraphics[width=1\linewidth]{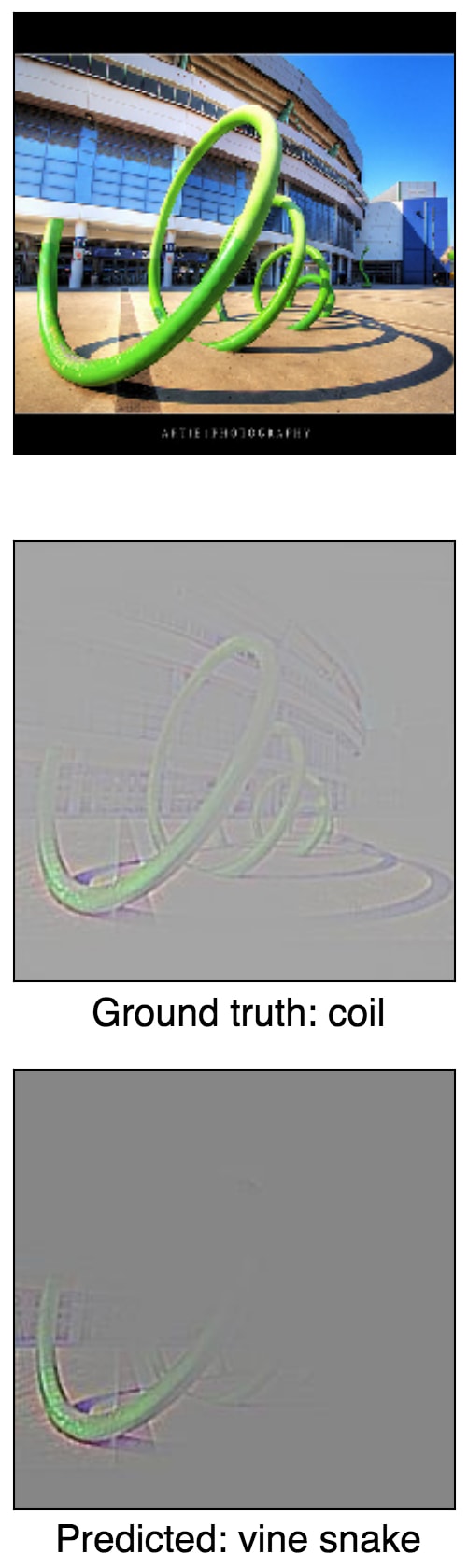}
        \vspace{-15pt}
        \caption{}
        \label{fig:failure_snake}
    \end{subfigure}
    \vspace{-4pt}
    \caption{\scriptsize{In these cases the model (VGG-16) failed to predict the correct class as its top 1
            prediction, but it even failed to predict the correct class in its top 5 for figure b.
            All of these errors are due in part to class ambiguity.
            (a) For example, the network predicts `sandbar' based on the foreground of (a), but it also knows where the correct label `volcano' is located.
        (c-d) In other cases, these errors are still reasonable but not immediately apparent.
        For example, humans would find it hard to explain the predicted `syringes' in (c) without looking at the visualization for the predicted class.
    }}
    \label{fig:failures}
    \end{center}
\vspace{-30pt}
\end{figure}

We use \cgb{} to analyze failure modes of the VGG-16 CNN on ImageNet classification~\cite{imagenet_cvpr09}.

In order to see what mistakes a network is making we first get a list of examples that the network (VGG-16) fails to classify correctly.
For the misclassified examples, we use \cgb{} to visualize both the correct class and the predicted class.
A major advantage of \cgb{} over other methods is its ability to more usefully investigate and explain classification mistakes, since our visualizations are high-resolution and more class-discriminative.
As seen in \figref{fig:failures}, some failures are due to ambiguities inherent in ImageNet classification.
We can also see that seemingly unreasonable predictions have reasonable explanations, which is a similar
observation to HOGgles~\cite{vondrick_iccv13}.


%% file: sections/experiments.tex
\vspace{-13pt}
\subsection{Image Captioning and VQA}

\vspace{-10pt}
\begin{figure}[ht!]
    \centering
  \begin{subfigure}[t]{0.49\textwidth}
    \includegraphics[width=\textwidth]{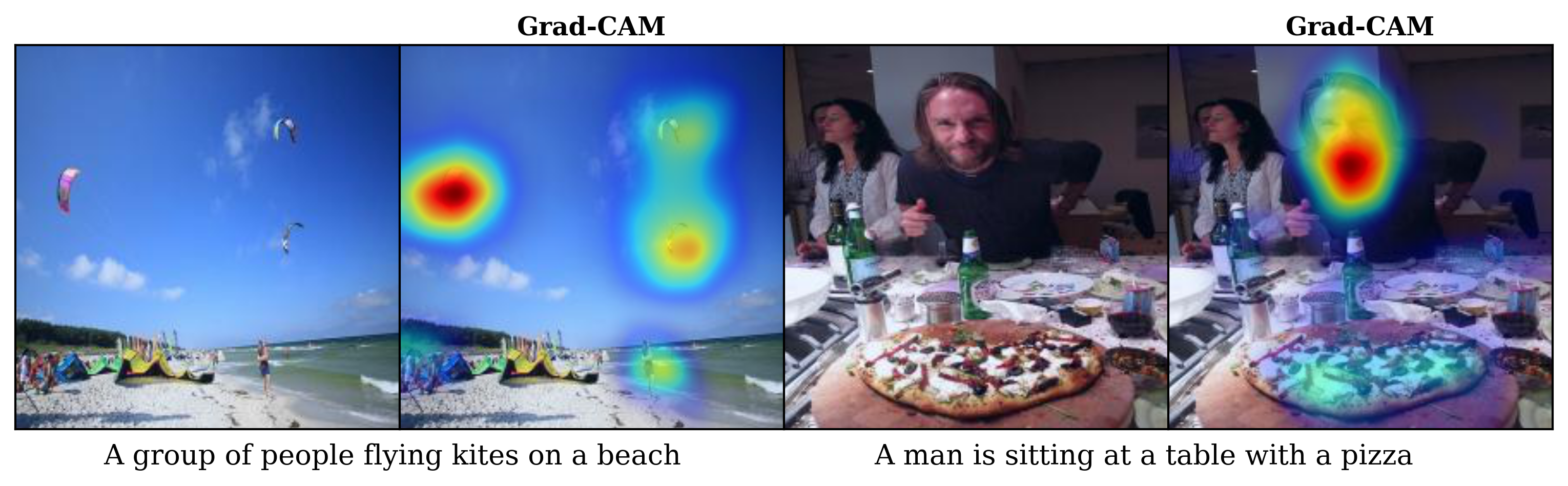}
    \vspace{-15pt}
    \caption{\scriptsize{\rp{Image captioning explanations}}}
  \label{fig:captioning}
  \end{subfigure}
  ~
    \begin{subfigure}[t]{0.49\textwidth}
    \includegraphics[width=\textwidth]{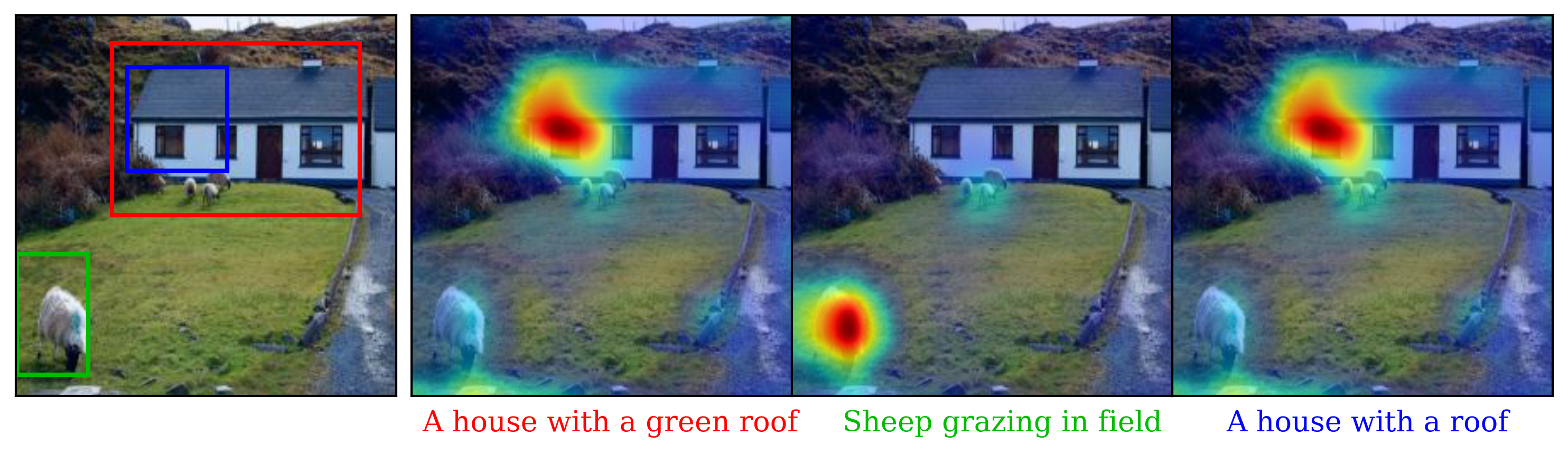}
    \vspace{-15pt}
    \caption{\rp{\scriptsize{Comparison to DenseCap}}}
   \label{fig:densecap}
   \end{subfigure}
    \vspace{-8pt}
    \caption{\scriptsize{Interpreting image captioning models: We use our class-discriminative localization technique, \gcam{} to find spatial support regions for captions in images. \reffig{fig:captioning} Visual explanations from image captioning model~\cite{karpathy2015deep} highlighting image regions considered to be important for producing the captions. \reffig{fig:densecap} Grad-CAM localizations of a \emph{global} or \emph{holistic} captioning model for captions generated by a dense
    captioning model~\cite{johnson_cvpr16} for the three bounding box proposals marked on the left. We can see that we get back \gcam{} localizations (right) that agree with those bounding boxes -- even though the captioning model and Grad-CAM do not use any bounding box annotations.}}

\vspace{-8pt}
\end{figure}

\para{Image Captioning}
We visualize spatial support for a simple image captioning model~\cite{karpathy2015deep}
\footnote{`neuraltalk2', publicly available at \url{https://github.com/karpathy/neuraltalk2}}
(without attention) using \gcam{} visualizations.
Given a caption, we compute the gradient of its log probability~\wrt units in the last convolutional layer of the CNN ($conv5\_3$ for VGG-16) and generate \gcam{} visualizations as described in \secref{sec:approach}.
Results are shown in \reffig{fig:captioning}.
In the first example, the \gcam{} maps for the generated caption localize every occurrence of both the kites and people in spite of their relatively small size.
In the top right example, \gcam{} correctly highlights the pizza and the man, but ignores the woman nearby, since `woman' is not mentioned in the caption.
As described in \reffig{fig:densecap}, if we generate captions for specific bounding boxes in an image then \gcam{} highlights only regions within those bounding boxes.

\para{Visual Question Answering}
Typical VQA pipelines 
consist of a CNN to model images and an RNN language model for questions.
The image and the question representations are fused to predict the answer, typically with a 1000-way classification. 
\begin{figure}[ht!]
    \vspace{-5pt}
    \begin{center}
    \begin{subfigure}[t]{1.0\columnwidth}
    \begin{center}
    \includegraphics[scale=0.23]{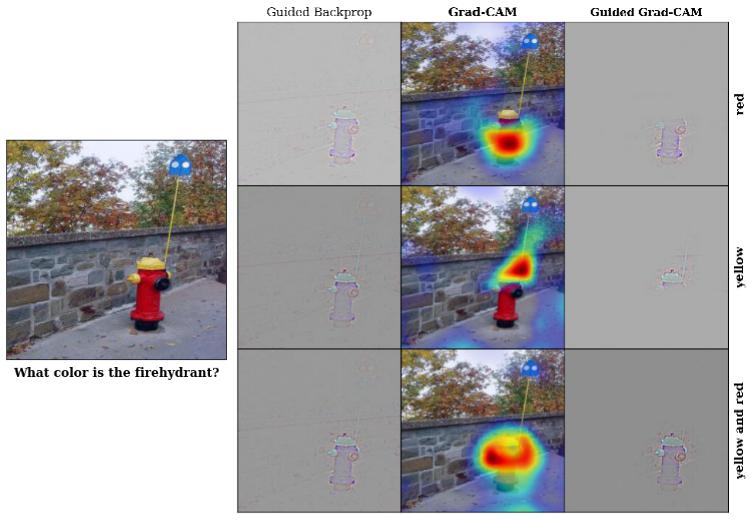}
\end{center}%
    \vspace{-12pt}
    \caption{\scriptsize{Visualizing baseline VQA model from \cite{Lu2015}}}
    \label{fig:vqa_main_fig}
    \end{subfigure}
    \begin{subfigure}[t]{1.0\columnwidth}
    \begin{center}
    \includegraphics[scale=0.22]{figures/vqa_residual_main.jpg}
\end{center}
    \vspace{-12pt}
    \caption{\scriptsize{Visualizing ResNet based Hierarchical co-attention VQA model from \cite{Lu2016}}}
        \label{fig:vqa_residual_main}
    \end{subfigure}
\end{center}
    \vspace{-7pt}
    \caption{\scriptsize{VQA visualizations: (a) Given the image on the left and the question ``What color is the firehydrant?'', we visualize \gcam{}s and \cgb{}s for the answers ``red", ``yellow" and ``yellow and red".
        Our visualizations are highly interpretable and help explain the model's predictions -- for ``red'', the model focuses on the bottom red part of the firehydrant; when forced to answer ``yellow'', the model concentrates on it`s top yellow cap, and when forced to answer ``yellow and red", it looks at the whole firehydrant! \rp{ (b) \gcam{} for ResNet-based VQA model.}}
    }
	\label{fig:vqa}
    \vspace{-8pt}
\end{figure}
Since this is a classification problem, we pick an answer (the score $y_c$ in \eqref{eq:scores}) and use its score to compute \gcam{}. 
Despite the complexity of the task, involving both visual and language components,
the explanations (of the baseline VQA model from \cite{Lu2015} and a ResNet based hierarchical co-attention model from \cite{Lu2016}) described in \reffig{fig:vqa}
are suprisingly intuitive and informative.

%% file: sections/conclusions.tex
\vspace{-5pt}
\section{Conclusion}
\vspace{-10pt}
In this work, we proposed a novel class-discriminative localization technique 
-- Gradient-weighted Class Activation Mapping (\gcam{}) -- 
and combined it with existing high-resolution visualizations 
to produce visual explanations for CNN-based models.
Human studies reveal that our localization-augmented visualizations can discriminate between classes more accurately and better reveal the trustworthiness of a classifier.
In addition to the image captioning and VQA examples shown here,
the full version of our paper~\cite{gradcam_arxiv} evaluates \gcam{} on the ImageNet localization
challenge, analyzes failure modes of VGG-16 on ImageNet classification using \gcam{},
measures correlation between VQA \gcam{} and human attention maps, describes ablation studies and provides many more examples.
Grad-CAM provides a new way to understand any CNN-based model.

\vspace{-10pt}

\vspace{-10.5pt}

%% file: main.bbl
\begin{thebibliography}{10}\itemsep=-1pt

\bibitem{agrawal2015cloudcv}
H.~Agrawal, C.~S. Mathialagan, Y.~Goyal, N.~Chavali, P.~Banik, A.~Mohapatra,
  A.~Osman, and D.~Batra.
\newblock {CloudCV: Large Scale Distributed Computer Vision as a Cloud
  Service}.
\newblock In {\em Mobile Cloud Visual Media Computing}, pages 265--290.
  Springer, 2015.

\bibitem{imagenet_cvpr09}
J.~Deng, W.~Dong, R.~Socher, L.-J. Li, K.~Li, and L.~Fei-Fei.
\newblock {ImageNet: A Large-Scale Hierarchical Image Database}.
\newblock In {\em CVPR}, 2009.

\bibitem{johnson_cvpr16}
J.~Johnson, A.~Karpathy, and L.~Fei-Fei.
\newblock {DenseCap: Fully Convolutional Localization Networks for Dense
  Captioning}.
\newblock In {\em CVPR}, 2016.

\bibitem{karpathy2015deep}
A.~Karpathy and L.~Fei-Fei.
\newblock Deep visual-semantic alignments for generating image descriptions.
\newblock In {\em CVPR}, 2015.

\bibitem{Lu2015}
J.~Lu, X.~Lin, D.~Batra, and D.~Parikh.
\newblock {Deeper LSTM and normalized CNN Visual Question Answering model}.
\newblock \url{https://github.com/VT-vision-lab/VQA_LSTM_CNN}, 2015.

\bibitem{Lu2016}
J.~Lu, J.~Yang, D.~Batra, and D.~Parikh.
\newblock {Hierarchical Question-Image Co-Attention for Visual Question
  Answering}.
\newblock In {\em NIPS}, 2016.

\bibitem{lime_sigkdd16}
M.~T. Ribeiro, S.~Singh, and C.~Guestrin.
\newblock {"Why Should I Trust You?": Explaining the Predictions of Any
  Classifier}.
\newblock In {\em SIGKDD}, 2016.

\bibitem{gradcam_arxiv}
R.~Selvaraju, A.~Das, R.~Vedantam, M.~Cogswell, D.~Parikh, and D.~Batra.
\newblock {Grad-CAM: Why did you say that? Visual Explanations from Deep
  Networks via Gradient-based Localization}.
\newblock {\em CoRR}, abs/1610.02391, 2016.

\bibitem{springenberg_arxiv14}
J.~T. Springenberg, A.~Dosovitskiy, T.~Brox, and M.~A. Riedmiller.
\newblock {Striving for Simplicity: The All Convolutional Net}.
\newblock {\em CoRR}, abs/1412.6806, 2014.

\bibitem{vondrick_iccv13}
C.~Vondrick, A.~Khosla, T.~Malisiewicz, and A.~Torralba.
\newblock {HOGgles: Visualizing Object Detection Features}.
\newblock {\em ICCV}, 2013.

\bibitem{zeiler_eccv14}
M.~D. Zeiler and R.~Fergus.
\newblock {Visualizing and understanding convolutional networks}.
\newblock In {\em ECCV}, 2014.

\bibitem{zhou_cvpr16}
B.~Zhou, A.~Khosla, L.~A., A.~Oliva, and A.~Torralba.
\newblock {Learning Deep Features for Discriminative Localization.}
\newblock In {\em CVPR}, 2016.

\end{thebibliography}
